\pdfoutput=1
\documentclass[11pt]{article}

\usepackage[final]{acl}
\usepackage{amsmath}
\usepackage{amssymb}
\usepackage{amsfonts}
\usepackage{times}
\usepackage{latexsym}
\usepackage{multirow}
\usepackage{bm}
\usepackage{bbm}

\usepackage{makecell}
\usepackage[T1]{fontenc}
\usepackage[utf8]{inputenc}

\usepackage{microtype}

\usepackage{inconsolata}

\usepackage{pifont}
\usepackage{booktabs}
\usepackage{enumitem}
\usepackage{tabularray}
\definecolor{LightBlue}{RGB}{212, 230, 246} % Example
\usepackage{graphicx}

\def\authnotes{1}
\newcounter{mynote}[section]
\newcommand{\notecolor}{blue}
\newcommand{\thenote}{\thesection.\arabic{mynote}}
\newcommand{\qnote}[1]{\ifnum\authnotes=1\refstepcounter{mynote}{\bf
    \textcolor{\notecolor}{$\ll$QL~\thenote: {\sf #1}$\gg$}}\fi}

\title{CR-UTP: Certified Robustness against Universal Text Perturbations on Large Language Models}

\author{Qian Lou\textsuperscript{1$+$},
  Xin Liang\textsuperscript{1$*$},
   Jiaqi Xue\textsuperscript{1$*$}, 
    Yancheng Zhang\textsuperscript{1}, 
  Rui Xie\textsuperscript{2},
Mengxin Zheng\textsuperscript{1}\\
  \textsuperscript{1}Department of Computer Science, University of Central Florida \\
    \textsuperscript{2}Department of Statistics and Data Science, University of Central Florida}

\begin{document}
\maketitle
\def\thefootnote{+}\footnotetext{Corresponding author: Qian Lou, \url{qian.lou@ucf.edu}}
\def\thefootnote{*}\footnotetext{These authors contributed equally to this work. }

\begin{abstract}
It is imperative to ensure the stability of every prediction made by a language model; that is, a language’s prediction should remain consistent despite minor input variations, like word substitutions. In this paper, we investigate the problem of certifying a language model’s robustness against Universal Text Perturbations (UTPs), which have been widely used in universal adversarial attacks and backdoor attacks.  Existing certified robustness based on random smoothing has shown considerable promise in certifying the input-specific text perturbations (ISTPs), operating under the assumption that any random alteration of a sample’s clean or adversarial words would negate the impact of sample-wise perturbations. However, with UTPs, masking only the adversarial words can eliminate the attack. A naive method is to simply increase the masking ratio and the likelihood of masking attack tokens, but it leads to a significant reduction in both certified accuracy and the certified radius due to input corruption by extensive masking. To solve this challenge, we introduce a novel approach, the \textit{superior prompt search} method, designed to identify a \textit{superior prompt} that maintains higher certified accuracy under extensive masking. Additionally, we theoretically motivate why ensembles are a particularly suitable choice as base prompts for random smoothing. The method is denoted by \textit{superior prompt ensembling} technique. We also empirically confirm this technique, obtaining state-of-the-art results in multiple settings. These methodologies, for the first time, enable high certified accuracy against both UTPs and ISTPs. The source code of CR-UTP is available at \url{https://github.com/UCF-ML-Research/CR-UTP }.
\end{abstract}

\section{Introduction}
% p1: Prompt-based language models (PLMs) have achieved significant success across a wide range of real-world applications. PLM's vulnerability to noisy inputs has notably restricted their utility, especially in high-stake environments. In such settings, it’s imperative to ensure the stability of every prediction made by PLMs; that is, PLM predictions should remain consistent despite minor input variations, like word substitutions. This concern aligns with the study of certified robust PLMs, which guarantees that all PLM predictions are accurate within a local vicinity of the input.

Prompt-based Language Models (PLMs)~\cite{thoppilan2022lamda, zeng2022glm, achiam2023gpt,touvron2023llama2, chiang2023vicuna} have achieved significant success across a wide range of real-world applications~\cite{wu2020app0, brown2020lapp1, wei2022app2, chowdhery2023app3}. However, despite their prominent performance, PLMs have been shown vulnerable to noises and perturbations on the input~\cite{xu2022vul, shayegani2023vul2, lou2022trojtext,al2023trojbits, zheng2023trojfsp}. Such vulnerability has notably restricted PLM's utility, especially in high-stake environments such as bank records analysis~\cite{heaton2017finance}, health care records analysis~\cite{myszczynska2020medical}. In these settings, the stability of every prediction is critical, i.e., PLM predictions should remain consistent despite minor input variations, such as word substitutions~\cite{alzantot2018wordper1, ren2019wordper2, li2020bertattack}. This concern aligns with the study of certified robust PLMs~\cite{zeng2023certified}, which guarantees that all PLM predictions are accurate within the local vicinity of the input.

% P2: What are the UTPs. Its difference against with ISTPs. Why UTPs are also important.
% \qnote{Add citations to UTPs and ISTPs, respectively. Explain the reasons why they are much stronger to defend and are harder to mitigate. Provide these attack examples. Add our TrojLLM citation too.}

Input perturbations can be classified into Universal Text Perturbations (UTPs) and Input-Specific Text Perturbations (ISTPs). UTPs are characterized by their ability to be applied across different inputs, making them transferable, whereas ISTPs are tailored to specific inputs. In detail, attackers employing ISTP strategies, exemplified by TextFooler~\cite{jin2020bert} and DeepWordBug~\cite{gao2018black}, craft a unique adversarial sentence for each target input sentence. Conversely, attackers utilizing UTP methodologies, such as those found in TrojLLM~\cite{xue2023trojllm} and UAT~\cite{wallace2019universal}, identify a single or a small number of adversarial tokens that can be inserted into any sentence to influence the model's prediction. This makes UTPs a more considerable threat to the robustness of PLMs since a specific set of adversarial tokens could lead to mispredictions across any input. Additionally, UTPs pose a greater challenge in mitigation compared to ISTPs. This challenge arises because ISTPs depend on weaker adversarial patterns that can be addressed by introducing modifications to the adversarial or clean tokens. However, UTPs are based on stronger adversarial patterns, which require exact identification and masking of the adversarial tokens for effective mitigation. %In practical scenarios, pinpointing the precise locations of adversarial tokens is often unfeasible, making the mitigation of UTPs a notably difficult task.

% P3: Existing certified robustness based on random smoothing has shown considerable promise in certifying the input-specific text perturbations (ISTPs), operating under the assumption that any random alteration of a sample’s clean or adversarial words would negate the impact of sample-wise perturbations. However, with UTPs, masking only the adversarial words can eliminate the attack.

% \qnote{highlight again that ISTPs can be mitigated by masking any adversarial perturbed words or clean words. Highlight this masking is easy but masking the adversarial perturbed words only is difficult since the location of the perturbed word is unknown. Only a large ratio can potentially mask this word with a large probability.}
Random smoothing has been recognized as an effective defense offering certified robustness for models in computer vision~\cite{horvath2021boosting-ensmeble} and NLP~\cite{zeng2023certified}, yet its application has been limited to ISTPs. This method assumes that random alterations to a sample's words counteract perturbations. However, this approach falls short against UTPs, which require precise masking of adversarial tokens for mitigation, unlike ISTPs which can be mitigated by randomly masking any tokens. Defending against UTPs is challenging due to the unknown positions of adversarial tokens, requiring a high mask ratio that could degrade PLM accuracy. Thus, ensuring certified robustness against UTPs in PLMs remains an unresolved challenge.

% P4: A naive method is to simply increase the masking ratio and the likelihood of masking attack tokens, but it leads to a significant reduction in both certified accuracy and the certified radius, due to input corruption by extensive masking.
% \qnote{add the fact that the defense effect against UTPs is also not significant. Try to use several sentences to explain the reasons. May use the logic chain that the ASR is heavily related to accuracy mentioned earlier.}
Naively increasing the masking ratio can improve the chances of covering adversarial tokens in UTPs, potentially reducing the Attack Success Rate (ASR). However, this method often results in only minor ASR improvements due to the trade-off with certified accuracy. High masking ratios in random smoothing significantly diminish certified accuracy, leading to randomized model inferences as a large portion of input tokens are obscured, leaving insufficient data for accurate predictions.

In this paper, we introduce CR-UTP, a method to equip PLMs with certified robustness against UTPs, achieving both high certified accuracy and low ASR. Our contributions are as follows:
\begin{itemize}
    \item We adapt the certified robustness method to PLMs and propose \textit{Superior Prompt Search} for robust prompts with masked inputs. 
    
    \item We introduce a prompt ensemble method to reduce the variance of masked inputs and increase the certified accuracy with theoretical analysis and empirical implementation.
 
    \item Through extensive experimentation, we show our CR-UTP effectively increases the certified accuracy by $\sim15\%$ and decreases the ASR by $\sim35\%$ compared to prior works.
\end{itemize}

\section{Background and Motivation}
\begin{figure*}
    \centering
    \includegraphics[width=0.95\linewidth]{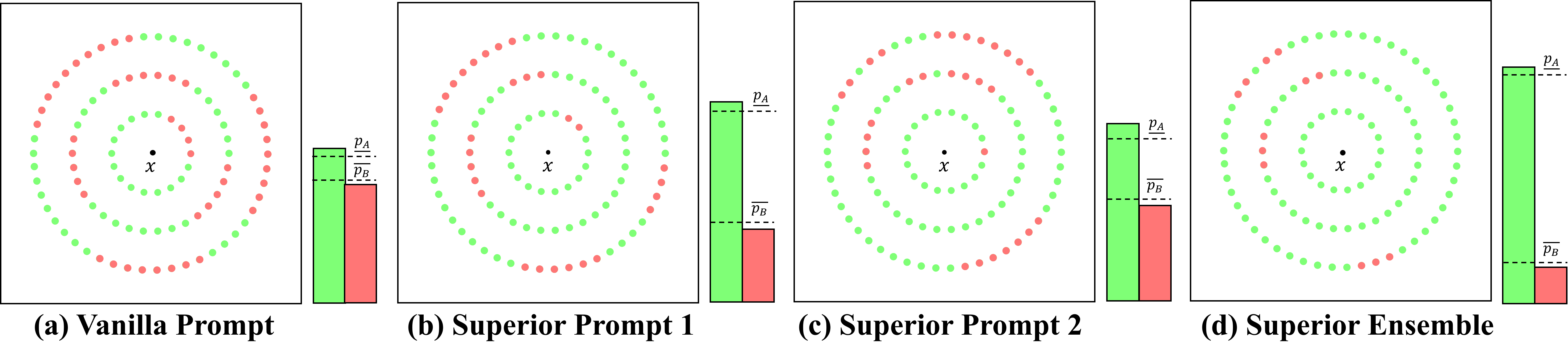}
    \caption{Illustration of the prediction distributions. A superior prompt exhibits greater robustness compared to a vanilla prompt, with ensembled prompts showing even higher robustness.
Different colors represent various classes, and different radii indicate varying levels of perturbation. The bars demonstrate the output class probabilities for the smoothed PLMs given corresponding prompts. 
$\underline{p_A}$ represents the minimum probability of the majority class, and $\overline{p_B}$ indicates the maximum probability of the second-most likely class.  }
    \label{fig:motivation}
\end{figure*}

\noindent\textbf{Adversarial Attacks.} Adversarial attacks in deep learning involve embedding a trigger into certain training samples, thereby creating poisoned datasets. When a deep learning model is trained on such tainted datasets, it behaves normally when presented with clean inputs but exhibits malicious behavior when encountering inputs containing the trigger. In the realm of visual images, triggers often manifest as tiny patches~\cite{zheng2023trojfair} or global perturbations~\cite{CTRL,zheng2023trojvit}. In the context of language data, triggers can be rare words or characters like "cf"~\cite{kurita2020weight}. This paper specifically focuses on text-based attacks.

\noindent\textbf{Text Adversarial Attacks.} Text adversarial attack methods generate adversarial sentences by perturbing original sentences to maximally increase the model's prediction error, while maintaining the fluency and naturalness of the adversarial examples. These attacks on prompt-based language models can be categorized into two groups: input-specific text perturbation attacks (ISTPs) and universal adversarial perturbation attacks (UTPs). In ISTP attacks, the attacker optimizes an adversarial sentence for each input, mainly by replacing, scrambling, and erasing characters (e.g., DeepWordBug~\cite{gao2018black} and HotFlip~\cite{ebrahimi2018hotflip}) or words (e.g., TextFooler~\cite{jin2020bert}). Conversely, UTP attacks optimize a universal trigger for a prompt-based language model, and the output of any input embedded with this trigger will be manipulated (e.g., TrojLLM~\cite{xue2023trojllm} and UAT~\cite{wallace2019universal}).

\noindent\textbf{Certified Robustness in Language Models.} 
Numerous defense methods, such as adversarial training~\cite{yoo2021towards},  model detection~\cite{zheng2023ssl} and perturbation-controlled approaches, have been developed to counteract adversarial attacks \cite{DBLP:journals/corr/abs-1909-06723,zhou2021defense,goyal2023survey}. However, these traditional tools may become ineffective against novel, advanced attack strategies. To address this, certified robustness has been introduced, offering a guarantee against any attack as long as the number of perturbed words remains below a certain threshold. A model achieves certification if it can consistently produce the correct output when the number of perturbations does not exceed the certified radius. While models of smaller size can obtain robustness certification through deterministic methods \cite{DBLP:journals/corr/abs-2009-04131,DBLP:journals/corr/abs-2201-01978,weng2018fast,DBLP:journals/corr/abs-1711-00851}, the computational demands of language models preclude such approaches. Consequently, probabilistic methods, such as Random Smoothing~\cite{pmlr-v97-cohen19c}, have been introduced to certify the robustness of large language models.

\noindent\textbf{Random Smoothing.} Random Smoothing, a promising approach introduced by \cite{pmlr-v97-cohen19c, DBLP:journals/corr/abs-2003-08904}, certifies the robustness of large neural networks. This method enhances a model's robustness by adding Gaussian noise to the original input \cite{DBLP:journals/corr/abs-2003-01908, li2023double}. It was quickly adopted for large language models in the NLP field, exemplified by SAFER \cite{Ye2020SAFERAS} and Randomized [MASK] \cite{zeng2023certified}. To improve model performance with random smoothing, the computer vision field has explored re-training the original model to adapt to Gaussian noise \cite{DBLP:journals/corr/abs-2006-04062, DBLP:journals/corr/abs-2001-02378,DBLP:journals/corr/abs-1906-04584}. However, applying this re-training method in NLP to achieve a model tolerant to smoothing is prohibitively expensive. Instead, in the NLP domain, \cite{zhang2023certified} proposed a self de-noising method that allows the Pretrained Language Model (PLM) itself to recover the information lost due to masking.

\begin{figure}[ht!]
    \centering
    \includegraphics[width=1\linewidth]{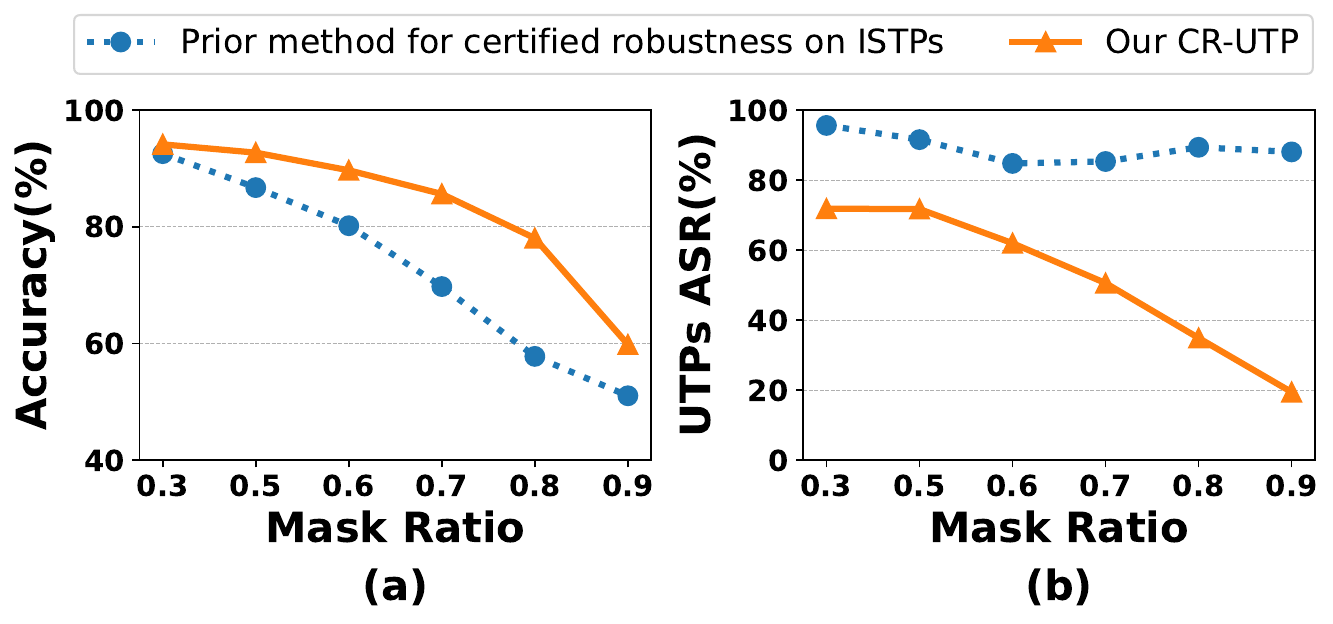}
    \caption{(a) Our CR-UTP shows higher certified robustness accuracy and (b) Our CR-UTP significantly reduces ASR.}
    \label{fig:motivation1}
\end{figure}

\begin{figure*}
    \centering
\includegraphics[width=1\linewidth]{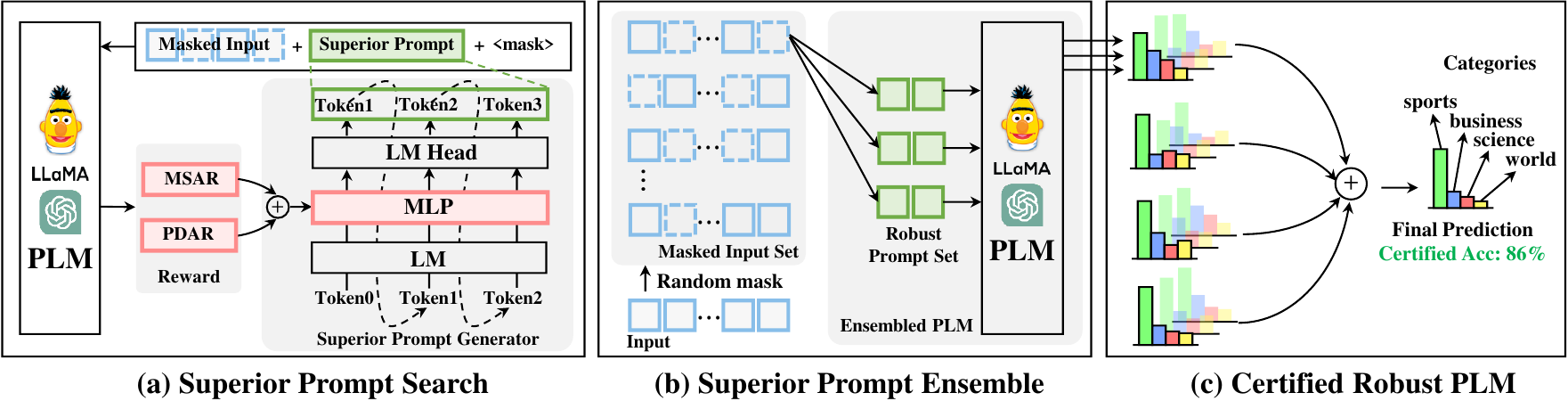}
    \caption{Overview of CR-UTP. CR-UTP leverages superior prompt search and prompt ensembling techniques to enhance the certified robustness of PLMs.}
    \label{fig:overview}
\end{figure*}

\noindent\textbf{Motivation.}
%As is shown in Figure \ref{fig:motivation1}, although prior works~\cite{zeng2023certified, zhang2023certified} leverage random smoothing to provide certified robustness against ISTPs, these methods cannot directly apply to UTPs. As is depicted in Figure \ref{fig:motivation1} (b), when the mask ratio is low, the ASR of UTPs in prior works is rather high, e.g., $\sim96\%$, in spite of the relatively high accuracy. While enlarging the mask ratio degrades the accuracy dramatically, the ASR can only be improved marginally. For example, when the mask ratio increases from $0.3$ to $0.8$, the ASR is only improved by $\sim6\%$, yet the accuracy decreases by $\sim35\%$. These empirical results indicate that prior ISTPs-targeted works can barely be applied to UTPs and motivate us to explore new designs to provide certified robustness against UTPs. By integrating superior prompt search and ensembles, we achieve stronger certified robustness against UTPs. Specifically, when the mask ratio increases from $0.3$ to $0.8$, we achieve $\sim6\times$ improvement on the ASR with $\sim2\times$ less accuracy drop compared with prior ISTPs-targeted works. We detail our UTPs-oriented prompt designs as follows.
Figure \ref{fig:motivation1} illustrates that while random smoothing has been effective for ISTPs as shown in previous works~\cite{zeng2023certified, zhang2023certified}, it struggles with UTPs. With a low mask ratio, the ASR for UTPs is high ($\sim$96\%), and increasing the mask ratio only marginally reduces ASR but significantly lowers accuracy. For example, increasing the mask ratio from 0.3 to 0.8 only reduces ASR by $\sim$6\% while accuracy drops by $\sim$35\%. These findings highlight the inadequacy of ISTP methods for UTPs and lead us to develop new techniques that combine superior prompt search and ensembles, significantly improving robustness against UTPs with less impact on accuracy. 

% (or use percentage, prior work ASR drops 6%, acc drops 35%, ours asr drops 37%, acc drops 16.1%)

% When applying random smoothing to PLMs, the mask ratio is a critical hyper-parameter deciding the robustness of the resulting model. As is shown in Figure \ref{fig:motivation1} (b), a high mask ratio, e.g. more than $0.7$, is typically needed to guarantee a reasonably low ASR, e.g., below $50\%$. This is because high mask ratio enlarges the likelihood that the adversarial tokens will be masked. However, as is shown in Figure \ref{fig:motivation1} (a), naively increasing the mask ratio will significantly reduce the model's accuracy while gaining only marginal improvement on the ASR. For example, when the mask ratio increases from $0.3$ to $0.7$, the ASR only decrease by less than $10\%$ while the accuracy decrease by $\sim25\%$. Accordingly, only increasing the mask ratios appears to be an inferior solution for defending against UTPs. Our primary results show that with proper design, it is possible to reduce ASR by over $20\%$ with only less than $10\%$ accuracy drop. This can be achieved thorough proper prompt design and prompt ensembles.

To achieve a low Attack Success Rate (ASR) against Universal Text Perturbations (UTPs), a high mask ratio, exceeding 0.5, is necessary. Yet, as Figure \ref{fig:motivation} (a) reveals, a vanilla prompt at this high mask ratio results in reduced accuracy due to the extensive masking of input tokens, which leaves limited information for precise classification. Figures \ref{fig:motivation} (b) and (c) illustrate that superior prompts can maintain higher accuracy under such conditions by incorporating random masking during the prompt design phase. This approach allows superior prompts to adjust to specific mask ratios, improving the lower bound ($\underline{p_A}$) on the majority class probability and reducing the upper bound ($\overline{p_B}$) on the runner-up class probability. However, these prompts still exhibit high variance across input samples. Ensembling techniques~\cite{liu2020ensemble, horvath2021boosting-ensmeble} reduce this variance, thereby enhancing robustness. As shown in Figure \ref{fig:motivation} (d), by combining superior prompts, we can leverage their individual advantages for a more favorable accuracy-ASR balance. This insight leads us to further investigate superior prompt design and ensembling as methods to bolster PLMs' certified accuracy and robustness against UTPs, aiming to lower the ASR while preserving high accuracy.

\section{CR-UTP Design}

\noindent\textbf{Overview.}
% challenge and method overview
%Prior works have employed random smoothing to effectively defense against LSTPs. While random smoothing can also apply to UTPs, a high mask ratio is typically needed to guarantee a reasonably low ASR. However, when a large fraction of input tokens are masked, the PLMs with vanilla prompt witness a severe drop in accuracy. We propose CR-UTP to provide PLMs with certified robustness against UTPs, which strikes a better balance between the accuracy and ASR.
In Figure \ref{fig:overview}, we detail the workflow of the proposed CR-UTP method. (a) Superior Prompt Search: we start with a basic prompt and employ a reinforcement learning approach to find a superior prompt adept at handling inputs with masked words. A unique reward function is utilized, which incorporates random masking during the prompt search phase to enhance the prompt's resilience to word masking. (b) Superior Prompt Ensemble: for making predictions, CR-UTP generates various versions of the original input by applying random masking. Each prompt assesses these versions and internally agrees on the optimal prediction. (c) Certified Robust PLM: CR-UTP aggregates the individual outcomes from each prompt through a second voting process to get the final, most robust prediction.

% Firstly, we generalize the random masking operation to binomial distribution and adapt the random masking operation to PLMs in Section~\ref{sec:method0}. Secondly, we introduce the superior prompt search in Section~\ref{sec:method1} to search for prompts that can achieve a high certified accuracy even when a large proportion of the input tokens are masked. This is achieved by introducing random mask in the prompt search phase, thus making the prompt mask-aware. By carefully designing the reward function, the proposed superior prompt search can achieve a high accuracy on masked input token sequence. In Section~\ref{sec:method2}, we proposed superior prompt ensemble to further improve the certified robustness. By integrating multiple prompts, prompt ensembling can effective reduce the variance of input samples, and thus enhancing the overall robustness of the PLMs.

In particular, we generalize the random masking operation to PLMs and analyze using random masking to defend UTPs in Section~\ref{sec:method0}. Also, we introduce the Superior Prompt Search in Section~\ref{sec:method1} to search for prompts that can achieve a high certified accuracy even when a large proportion of the input tokens are masked. Finally, in Section~\ref{sec:method2}, we proposed a Superior Prompt Ensemble to further improve the certified robustness. 

\subsection{Adapting Random Smoothing to PLMs}
\label{sec:method0}
Using  the random masking approach from Randomized [MASK] \cite{zeng2023certified}, the random masking operation $\mathcal{M}:\mathcal{X}\times \mathcal{M}(h,k)\rightarrow \mathcal{X}_{mask}$ would take an input text $\bm{x}=x_1x_2...x_h$ with $h$  words and randomly replacing $(h-k)$ word with the [MASK] to get the masked version $\mathcal{M}(\bm{x})$. Following this, we define a smoothed classifier $g(\bm{x})$ built upon the base classifier $f$ as follows:
\begin{equation}\label{eqn:random function}
\small
    g(\bm{x})=\mathop{\arg\max}\limits_{c\in \mathcal{C}}\Big[ \mathop{\mathbb{P}}\limits_{\mathcal{H} \sim  \mathcal{U}( h, k)}( f(\mathcal{M}(\bm{x}\mid \mathcal{H}))=c)\Big] 
  \end{equation}
  Then it can be shown that $g(x)$ would return $c$ when the certified condition is satisfied with probability at least $(1-\alpha)$ from Theorem 1 in \cite{zeng2023certified}.

  We define a prompt-based language model for classification tasks as ${f: \mathcal{X} \to \mathcal{C}}$, where $\mathcal{X}$ represents the domain of input texts and $\mathcal{C}={1,2,...,n_c}$ denotes the set of classification labels. The response of an input $\bm{x}$ to prompt $p$ is $y=f(p,\bm{x})$. The smoothed classifier $g$ over such PLM can be expressed as 
\begin{equation}\label{eqn:random function}
\small
    g_p(\bm{x})=\mathop{\arg\max}\limits_{c\in \mathcal{C}}\Big[ \mathop{\mathbb{P}}\limits_{\mathcal{H} \sim  \mathcal{H}( h, k)}( f(p,\mathcal{M}(\bm{x}\mid \mathcal{H}))=c)\Big] 
  \end{equation}
and the same certification can be achieved under such prompt-based model.

Defending against universal attacks with random masking necessitates a high mask ratio, especially in the context of UTPs, where the presence of any UTP token in the masked input guarantees the attack's success. Therefore, to ensure the adversarial token is masked with a probability exceeding 50\% for the smoothed function $g(\bm{x})$ to yield correct outcomes, the masking probability for each UTP token needs to be more than $0.5$. Consequently, for a UTP of length $r$, the likelihood that all UTP tokens are masked should be $p^r > 0.5$, implying $p > \sqrt[r]{0.5}$. For instance, $p > 0.5$ for $r=1$, and $p > 0.707$ for $r=2$. Ensuring correct results for inputs masked without UTP requires the model to perform effectively at high mask ratios. To alleviate the effects of extensive masking, we introduce a technique that enhances model performance under random masking without necessitating retraining of the language model, thereby reducing computational overhead.

\subsection{Superior Prompt Search}
\label{sec:method1}

Certified accuracy is influenced by the model's performance on randomly masked sentences, but high mask ratios can decrease accuracy due to loss of critical information. Enhancing a prompt-based language model's certified accuracy involves improving its tolerance to information loss from random masking. However, in few-shot and black-box scenarios, fine-tuning the pre-trained model or using gradient-based optimization for continuous prompts is infeasible. To tackle this challenge without gradient optimization, we approach it as a reinforcement learning (RL) problem to discover a discrete, robust prompt—termed a superior prompt. A direct approach involves searching for this prompt using datasets with randomly masked sentences to acclimate the model to diverse masking scenarios. Nonetheless, at high mask ratios (e.g., 70\%), the reduced information in few-shot datasets limits the effectiveness and generalizability of robustness enhancements. To overcome this, we suggest aligning the superior prompt's predictions on masked sentences with the vanilla prompt's on unmasked sentences, leveraging the existing knowledge of the vanilla prompt to offset the drawbacks of few-shot datasets and information loss from masking.

Our aim, as expressed in Equation~\ref{e:main_objective}, involves identifying an optimized prompt $p_s$ that augments a basic vanilla prompt $p_v$ by adding a sequence of $T$ tokens from the vocabulary $\mathcal{V}$. This strategy is intended to boost the smoothed function $g_{p_s}(\bm{x_i})$ 's accuracy on inputs $\bm{x_i}$ . The dataset $\mathcal{D}$ is composed of pairs of input sentences $x_i$ and their associated labels $y_i$.

% \begin{equation}
% %\footnotesize
% \max_{p_s\in \mathcal{V}^T}\sum_{(x_i,y_i)\in\mathcal{D}}\mathcal{ACC}(f(p_s, \hat{x_i}), y_i)
% \label{e:main_objective}
% \end{equation}

\begin{equation}
%\footnotesize
\max_{p_s\in \mathcal{V}^T}\sum_{(x_i,y_i)\in\mathcal{D}} \mathbb{I}(g_{p_s}(\bm{x_i})=y_i)
\label{e:main_objective}
\end{equation}

% However, the optimization objective presented in Equation~\ref{e:main_objective}, can prove to be intractable due to the discrete nature of tokens in \(p_s\), which are not conducive to gradient-based optimization. Furthermore, a black-box setting does not grant access to gradients. A brute-force search would have an exponential complexity of \(\mathcal{O(V^T)}\). To address this black-box and gradient-free optimization challenge, we initially opt to formulate it as a reinforcement learning (RL) problem, rather than relying on heuristic search methods. 

% \subsubsection{Masked Sentence Accuracy Reward}
\noindent\textbf{Masked Sentence Accuracy Reward.}
We introduce a two-fold reward function to guide the RL-based search for an optimal $p_s$. The first component, known as the Masked Sentence Accuracy Reward (MSAR), is designed to directly maximize the PLM's accuracy on masked sentences:

\begin{equation}
\small
\label{e:MSAR}
\mathcal{R}_{\texttt{MSAR}} = \sum_{(\bm{x_i},y_i)\in\mathcal{D}} \eta_1^{1-sign} \eta_2^{sign} Distance(\mathcal{M}(\bm{x_i}),y_i)
\end{equation}
where the \(Distance(\mathcal{M}(\bm{x_i}),y_i)\) denotes \(l_{y_i}(p_s,\mathcal{M}(\bm{x_i})) - max_{y'\neq y}l_{y'}(p_s,\mathcal{M}(\bm{x_i}))\), the difference of the correct logit and the maximum of the incorrect logits. The distance value is positive for correct predictions and negative otherwise. We denote the distance sign as \(sign=\mathbbm{1}[Distance(\mathcal{M}(\bm{x_i}),y_i) > 0]\). For a correct prediction (i.e., \(sign=1\)), we multiply the positive reward by a large number \(\eta_2\) to indicate its desirability; otherwise, we multiply the negative rewards by another number \(\eta_1\). This reward aims to maximize the PLM's accuracy on masked sentences.

% \subsubsection{Predictive Distribution Alignment Reward}
\noindent\textbf{Predictive Distribution Alignment Reward.}
To mitigate the challenge posed by information loss due to word masking, which is exacerbated in a few-shot setting, we propose an additional reward function, a.k.a, Predictive Distribution Alignment Reward (PDAR). It is designed to minimize the KL divergence between the predictive distributions of the vanilla prompt on unmasked sentences and the superior prompt on their masked equivalents:

%Nonetheless, directly formulating the problem as an RL problem to search for a superior prompt \(p_s\) using the masked sentence \(\hat{x}\) presents significant challenges. One primary issue is the loss of information due to word masking; for instance, a \(70\%\) mask ratio essentially obfuscates \(70\%\) of the tokens in a sentence with the \texttt{[MASK]} token. This reduction in available information is particularly problematic in a few-shot setting, where the model has limited examples to learn from. To address this, we introduce a novel reward function, \(\mathcal{R}_{KL}\), defined as:

\begin{equation}
\small
    \mathcal{R}_\texttt{PDAR} = -\sum_{(\bm{x_i},y_i)\in\mathcal{D}} KL(l(p_v,\bm{x_i}) \parallel l(p_s,\mathcal{M}(\bm{x_i})))
\end{equation}

This reward is designed to ensure that \(p_s\) retains alignment with \(p_v\)'s predictive behavior, thereby leveraging the foundational knowledge encoded in \(p_v\) to inform predictions in the face of partial information. Such strategic alignment enables \(p_s\) to infer missing data from the masked inputs, drawing on the robust insights and patterns encapsulated by \(p_v\). This method not only addresses the direct impact of masking on information availability but also enhances the model's capacity for generalization from limited examples.

% \subsubsection{Policy Model Update}
\noindent\textbf{Policy Model Update.}
As Figure~\ref{fig:overview} shows, the RL search process involves an agent that sequentially selects tokens \([s_1, ..., s_T]\) to construct the superior prompt \(p_s\), aiming to maximize the combined reward \(\mathcal{R} = \mathcal{R}_{\texttt{MSAR}} + \mathcal{R}_{\texttt{PDAR}}\). For each time step \(t\), the agent, given the previous tokens \(s_{<t}\), generates the next token \(s_t\) based on the policy generator \(G_{\theta_s}(s_t|s_{<t})\). Completion of \(p_s\) triggers the computation of the task reward \(\mathcal{R}\). To facilitate this, we employ a GPT-2 model as the backbone for our policy generator, enhanced with an insertable trainable Multilayer Perceptron (MLP) layer. The optimization focuses on the parameters of this MLP layer, tailoring the policy generator to effectively navigate the prompt construction space under the guidance of the designed reward function.

%The underlying reason is that, randomized masking adds noise to the input before model prediction, and its certification performance depends largely on the prompt-based language model performance on these corrupted data.

% During this process, an agent sequentially selects prompt tokens \([s_1, ..., s_T]\) to maxmize the reward formed in Equation 

% \begin{equation}
    
%     \label{e:reward}
% \end{equation}

% (Optional)P3: Analysis and summary (performance and overhead )

\subsection{Superior Prompt Ensemble}
\label{sec:method2}

%P1: Design principle and intuition

% P2: Why large masking ratio bring a large variance? certified accuracy decreased? Conclusion: A large variance introduces a large accuracy decrease. Consider adding a figure to show the increased variance with a larger mask ratio. 

% P5: Techs. The figure used for overview. Equations.  

% Why Ensembling prompt can reduce variance and increase certified accuracy. 

%p1: Overview of ensemble model. 

 Instead of relying on a single model, ensemble methods leverage the strengths and mitigate the weaknesses of various base classifiers. The core idea behind ensemble modelling is that a group of weak learners can come together to form a strong learner, thereby improving the model's ability to generalize well to unseen data. In the following section, we will demonstrate how random masking can increase the variance of the output probability and how model ensembling can mitigate the performance drop introduced by random masking.

%P2: Why large masking ratio bring a large variance? certified accuracy decreased? Conclusion: A large variance introduces a large accuracy decrease. Consider adding a figure to show the increased variance with a larger mask ratio. 

Suppose the probability that the smoothed model with prompt $p$ outputs the ground truth $y$ is $P^y(\bm{x}) = \mathop{\mathbb{P}}_{\mathcal{H} \sim  \mathcal{U}(h, k)}(f(p,\mathcal{M}(\bm{x}\mid \mathcal{H}))=y)$ in $[0,1]$, then the final probability output of the smoothed model is determined by two random variables, $P^y=P^y_o+P^y_m$ \cite{horvath2021boosting-ensmeble}, where $P_o^y$ is determined by the performance of the language model $f$ with prompt $p$ on the original input, and $P_m^y$ corresponds to the performance of prompt $p$ when the input $\bm{x}$ is randomly masked. Although $P_o$ should be constant 0 or 1 without any perturbation on the input, we could assume $P_o^y=l^y(p,\bm{x})$, the logit of the correct label under the random masking operation.
%Since $P_o$ is influenced by the performance of prompt $p$ and $P_m$ mostly depends on the perturbation ratio $\rho$ on the clean input $\bm{x}$, we can assume that $y_o$ and $y_m$ are independent. Suppose $P_o$ has mean $a_c$ and covariance $\sigma^2_c$, $P_m$ has mean $a_m$ and covariance $\sigma^2_m$. Under this assumption, the probability of the smoothing operation for each prompt $p$ has expectation and variance as follows:
% \[
% \mathbb{E}[P]=a_o + a_m,\quad  \sigma^2=\sigma^2_o + \sigma^2_m.
% \]
%Since the mask operation can cause information loss in $\bm{x}$, we assume $y_m$ has a negative mean $a_m$. 
We empirically analyse the performance of the mask operation variance $\sigma^2$ with respect to the perturbation rate, as shown in figure \ref{fig:res_acc}, and conclude that the perturbation ratio significantly influences variance. As the perturbation ratio increases, the variance initially rises and then decreases. This pattern occurs because, as the mask ratio increases from $0$ to $0.6$, the masking operation introduces more noise to the input, increasing the variance of $P^y$. However, when the mask ratio increases from $0.6$ to $1$, the remaining information decreases, leading the model to randomly guess any label, i.e., $P^y \rightarrow 1/n_c$, and the variance $\sum^y \rightarrow 0$ as more words in the sentence are masked.

%P3: Design principle. The model ensemble is not feasible. Not model ensemble, variance decrease? 
%P4: prompt ensembling. Advantages. 

With a high mask ratio, the random masking operation can significantly increase variance and reduce accuracy relative to the clean classifier's output. Model ensembles effectively decrease the overall variance of voting outcomes, thereby improving the likelihood of accurate predictions as the number of ensembles increases \cite{horvath2021boosting-ensmeble}. However, the high computational cost makes training multiple language models unfeasible. Consequently, we introduce a technique that ensembles a set of prompts during inference to exploit the distinctive feature of PLMs, wherein the initial prompt markedly affects the model's final output. By employing the superior prompt search method, we can create a collection of prompts that withstand the random masking operation, with the ensemble of these prompts' outputs demonstrating enhanced performance on heavily masked inputs.

Formally, we construct an ensemble classifier $\bar{f}$ with a set of $k$ different prompts $\bm{P}=\{p_i\mid i=1,..k\}$,  via hard voting of all the outputs from different prompts $p_i$,
\begin{equation}
\bar{f}(\bm{P}, \bm{x})=
\mathop{\arg\max}\limits_{c\in \mathcal{C}}\sum\limits_{i=1}^{k} \mathbb{I}(f(p_i, \bm{x})=c) 
\end{equation}
where $\mathbb{I}(f(p_i,\bm{x})=c)$ is the indicator function that equals 1 when $f(p_i, \bm{x})$ output $c$ and 0 otherwise. So the ensemble classifier would output the class that most of the prompts agree on.

Since the prompts ensemble operates as a single model, the certified robustness condition remains applicable to the assembled model $\bar{f}(\bm{P},\bm{x})$. Therefore, we can establish a new smoothing function $\bar{g}(\bm{x})$ by applying the same random masking operation to $\bm{x}$. Building on our previous findings, we anticipate performance improvements through the prompts ensemble. Our analysis of how the number of ensembles impacts the final probability outcome, as depicted in Figure~\ref{fig:res_ens_num}, demonstrates a substantial increase in the accuracy of the model ensemble with a concurrent reduction in variance as the number of ensembles increases.

%(Optional)P3: Analysis and summary (performance and overhead )

\section{Experimental Methodology}
\noindent\textbf{Datasets and Model.} 
In our evaluation, we utilize the SST-2 dataset~\cite{socher2013recursive} and Yelp~\cite{DBLP:journals/corr/Asghar16} for binary classification tasks, AgNews dataset~\cite{zhang2015character} for a four-class classification task. We adopt a 16-shot setting, which represents a typical few-shot scenario. Our experiments are mainly conducted with the widely-used pre-trained language model RoBERTa-large~\cite{liu2019roberta}, an advanced version of BERT~\cite{kenton2019bert} with 24 layers of Transformer architecture. We also evaluate the performance on large language models such as Llama2-7b~\cite{touvron2023llama} and GPT-3.5~\cite{brown2020language}. 

\noindent\textbf{Evaluation Metrics.} We adopted three key metrics in evaluations same with~\cite{zeng2023certified}. Clean accuracy \textbf{(CACC)} refers to the classification accuracy on clean sentences. The attack success rate \textbf{(ASR)} quantifies the percentage of input instances perturbed by an attack that successfully causes the model to make incorrect predictions. The poisoned accuracy \textbf{(PACC)} indicates the accuracy of the prompt-based language model on poisoned samples crafted from an attack.

%\begin{itemize}
   % \item The clean accuracy (CACC) is the classification accuracy on the clean sentences.
    %\item The attack success rate (ASR) quantifies the percentage of input instances perturbed by an attack that successfully causes the model to make wrong predictions.
    %\item The poisoned accuracy (PACC) denotes the accuracy of the prompt-based language model on poisoned samples crafted from an attack.
%\end{itemize}

\begin{table*}[th!]
\centering
\footnotesize
\setlength{\tabcolsep}{5pt}
\caption{Comparison of CR-UTP and Random Mask against attacks with a 70\% mask ratio on SST-2 dataset.}

\begin{tabular}
{lccccccccc}\toprule
\multirow{2}{*}{Scheme} & \multicolumn{3}{c}{w/o defence}& \multicolumn{3}{c}{Randomized [MASK]} & \multicolumn{3}{c}{CR-UTP} \\\cmidrule(lr){2-4}\cmidrule(lr){5-7}\cmidrule(lr){8-10}
  & CACC & ASR & PACC & CACC & ASR & PACC & CACC & ASR & PACC \\\midrule
DeepWordBug & 92.69 & 93.04 & 6.96 & $81.13$ & $45.18$ & $54.82$ & $85.61$ & $21.25$ & $78.75$  \\
TextFooler &  92.69 & 91.87 & 8.13 & $81.60$ & $42.88$ & $57.12$ & $85.28$ & $37.39$ & $62.61$ \\
UAT & 92.48 & 96.85 & 52.97 &  $80.75$ & $79.92$ & $60.18$ & $85.53$ & $50.63$ & $75.92$ \\
TrojLLM & 92.69 & 91.88 & 53.76 & $80.94$ & $85.31$ & $56.84$ & $85.70$ & $50.55$ & $73.04$ \\
\bottomrule
\end{tabular}
\label{t:comprasion_with_prior_work}
\end{table*}

\noindent\textbf{Evaluated Attacks.} We evaluated our CR-UTP under two input-specific text perturbation \textbf{(ISTP)} adversarial attacks, TextFooler~\cite{jin2020bert} and DeepWordBug~\cite{gao2018black}, and two universal text perturbation \textbf{(UTP)} attack, UAT~\cite{wallace2019universal} and TrojLLM~\cite{xue2023trojllm}. TextFooler adversarially perturbs the text inputs by the word-level substitutions, whereas DeepWordBug performs the character-level perturbations to each input by replacing, scrambling, and erasing a few characters of some words. The UAT attack generates universal adversarial triggers as sequences of tokens that are independent of input and, when appended to any dataset entry, prompt the model to generate a particular prediction. UTP attack TrojLLM uses reinforcement learning to search a universal trigger for a prompt-based language model, any text inputs with this trigger will lead to the model output target label.

\noindent\textbf{Implementation Details.} For the superior prompt generator configuration, we adhered to the parameters established in RL-Prompt~\cite{deng2022rlprompt}. Specifically, we use distilGPT-2, a large model with 82 million parameters, as a policy model for all tasks. Additionally, we use a multilayer perceptron (MLP) with one hidden layer which has 2,048 hidden states, added to distilGPT-2's existing 768 hidden states. For the hyperparameters of reward functions in the Equation~\ref{e:MSAR}, we set balancing weights $\eta_1=180$ and $\eta_2=200$. During inference of CR-UTP, we use an ensemble number of 5 with the best 5 prompts derived from the superior prompt search. During the certification process, the prediction number is 500 and the certification number is 1000. When using randomized [MASK] to defend against adversarial attacks, the voting number is set to 100. All experiments are conducted on a single Nvidia Geforce RTX-3090 GPU. Searching time for superior prompts on SST-2 is $3.8$ hours, the certification time for one sentence is $\sim 8$ seconds. Further details about training time and inference efficiency are provided in the appendix~\ref{sec:appendix}.

\label{sec:experiments}

\section{Results}

\subsection{Comparison of CR-UTP with Random Mask.} 
In Table~\ref{t:comprasion_with_prior_work}, we conducted experiment comparing the performance of CR-UTP with no defence input and Randomized [MASK]~\cite{zeng2023certified} at a $70\%$ masking ratio against two ISTP adversarial attacks, i.e.,  DeepWordBug~\cite{gao2018black}, TextFooler~\cite{jin2020bert}, and two UTP adversarial attacks, i.e., UAT~\cite{wallace2019universal} and TrojLLM~\cite{xue2023trojllm}. CR-UTP significantly reduces the ASR from 96.85\% to 50.63\% on UAT attack, and 93.04\% to 21.25\% on DeepWordBug attack, which suggests random masking operation with a high mask ratio could effectively reduce attack success rate on both ISTP and UTP adversarial attacks. Our CR-UTP exhibits superior performance over Randomized [MASK] across all metrics in the evaluated adversarial attacks. Furthermore, CR-UTP exhibits higher CACC and PACC than the original input and Randomized [MASK]. This improvement is attributed to its efficient prompt search method, which identifies robust prompts to random mask, and superior prompt ensemble technique, further reducing CACC variance. Moreover, CR-UTP achieves a substantial reduction in attack success rate (ASR), averaging a $21.4\%$ greater decrease compared to Randomized [MASK], with a remarkable $34.76\%$ ASR reduction in the TrojLLM attack. This enhancement stems from CR-UTP's ability to leverage the differential outputs of various prompts, enabling a robust ensemble prediction for improved defence outcomes against adversarial samples. Additionally, CR-UTP demonstrates a notable increase in poisoned accuracy (PACC), indicating its ability to maintain high accuracy even under attack scenarios.

\begin{table*}[ht!]
\centering
\footnotesize
\setlength{\tabcolsep}{5pt}
\caption{An ablation study of CR-UTP techniques. Our baseline is random mask with \(70\%\) ratio; \(\mathcal{R}_{\texttt{MSAR}}\) denotes employing superior prompt search only using reward \(\mathcal{R}_{\texttt{MSAR}}\); \(\mathcal{R}_{\texttt{MSAR}}+\mathcal{R}_{\texttt{PDAR}}\) means using superior prompt search with rewards \(\mathcal{R}_{\texttt{MSAR}}\) and \(\mathcal{R}_{\texttt{PDAR}}\); CR-UTP incorporates all proposed techniques.}
\begin{tabular}{lccccccccc}\toprule
\multirow{2}{*}{Dataset} & \multicolumn{3}{c}{SST-2} & \multicolumn{3}{c}{AgNews} & \multicolumn{3}{c}{Yelp}\\\cmidrule(lr){2-4}\cmidrule(lr){5-7}\cmidrule(lr){8-10}
 & CACC & ASR & PACC & CACC & ASR & PACC  & CACC & ASR & PACC\\\midrule
w/o defense & $92.69$ & $91.88$ & $53.76$ & $88.91$  & $94.54 $ & $78.64$ & $95.42$&	$87.90$&	
$51.42$\\
Our baseline & $70.50$ & $85.31$ & $56.84$  & $80.94$ & $22.42$ & $75.71$ & $76.10$&	$58.26$&	$68.35$ \\
\(\mathcal{R}_{\texttt{MSAR}}\) & $81.93$  & $47.28$ & $76.83$   & $82.09$ & $21.31$ & $75.78$ & $83.23$&	$22.93$&	$81.93$\\
\(\mathcal{R}_{\texttt{MSAR}}+\mathcal{R}_{\texttt{PDAR}}\) & $84.90$ &	$63.93$	& $66.61$ & $83.06$ & $18.89$ & $78.65$ & $83.84$&	$20.83$&	$82.52$ \\
CR-UTP & $85.70$ & $50.55$ & $73.04$  & $84.27$ & $18.82$ & $78.73$  & $85.62$ &	$20.05$ &	$82.85$
\\\bottomrule
\end{tabular}

\label{t:ablation_techniques}
\end{table*}

\subsection{Ablation Study}
In this section, we explore the design space of CR-UTP and study the impact of various settings of CR-UTP on its attacking effects using the RoBERTa-Large with SST-2 dataset.

\noindent\textbf{CR-UTP Techniques Performance.} In Table~\ref{t:ablation_techniques}, we analyze the impact of different CR-UTP techniques on performance against TrojLLM on the SST-2, AgNews 
 and Yelp-2 datasets. For SST-2, utilizing the adapted random mask method (our baseline) leads to a significant drop in CACC by over $22\%$, mainly due to the loss of information from masking $70\%$ of the words. However, incorporating superior prompt search with reward \(\mathcal{R}_{\texttt{MSAR}}\), improves CACC by $11\%$ as the superior prompt proves more robust to random masking. Furthermore, combining rewards \(\mathcal{R}_{\texttt{MSAR}}+\mathcal{R}_{\texttt{PDAR}}\), further increases CACC to $84.90\%$ by enhancing prompt search effectiveness with \(\mathcal{R}_{\texttt{PDAR}}\), which aligns outputs for clean and masked sentences. Finally, employing the superior prompt ensemble technique elevates CACC to $85.70\%$ and reduces ASR from $91.88\%$ to $50.55\%$, indicating significant improvements over the baseline method. Similarly, on AgNews dataset, CR-UTP surpasses the baseline with a $3.33\%$ increase in CACC to $84.27\%$ and a $3.6\%$ decrease in ASR, highlighting CR-UTP's effectiveness. 
 Our CR-UTP method is also effective on the longer datasets such as Yelp dataset, which enhances the CACC and the PACC by 9.52\% and 13.5\%, respectively. Additionally, it reduces the ASR by 10.35\%.

\begin{figure}[h!]
    \centering
    \includegraphics[width=1\linewidth]{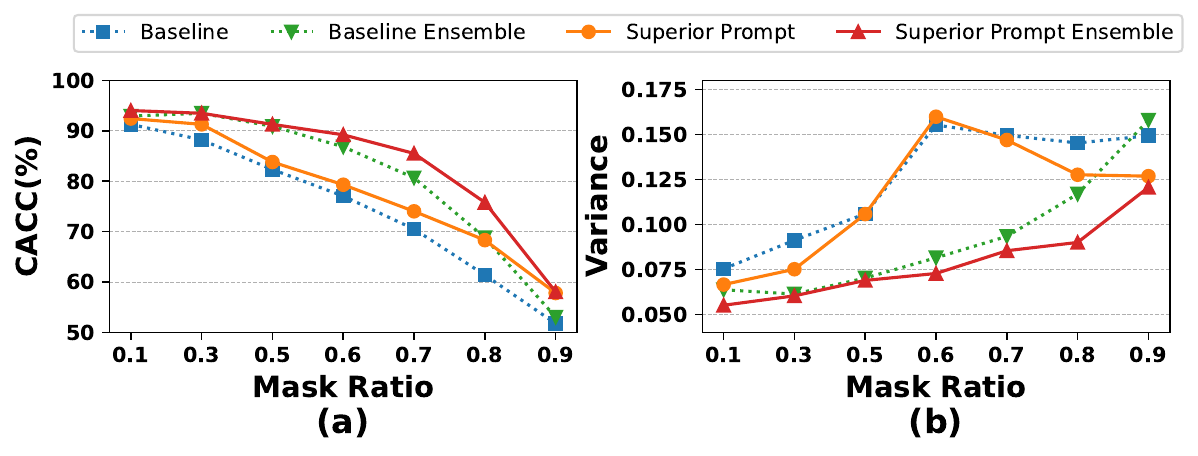}
    \caption{ (a) Clean accuracy and (b) variance of proposed methods under different mask ratio.}
    \label{fig:res_acc}
\end{figure}

\noindent\textbf{Mask Ratio.} To examine the effect of mask ratio on clean accuracy, we conduct experiments on the SST-2 dataset with varying mask ratios. Results in Figure~\ref{fig:res_acc} (a) show that while the baseline method's accuracy sharply drops from \(91.27\%\) to \(51.78\%\) as the mask ratio increases from \(10\%\) to \(90\%\), our superior prompt search technique leads to a more gradual decline, from \(92.42\%\) to \(57.82\%\). Additionally, employing our superior prompt ensemble method maintains a higher accuracy of \(85.70\%\) even at a \(70\%\) mask ratio, representing a significant improvement over the baseline method. 

In Figure~\ref{fig:res_acc} (b), the variance analysis of certified accuracy shows that while increasing the mask ratio results in higher variance for both baseline and superior prompt methods, the use of ensemble techniques, particularly the superior prompt ensemble method, reduces variances, providing a more consistent output despite the effects of masking. The variance peaks at the \(60\%\) mask ratio, indicating the highest sentence diversity. This suggests that the variance is influenced not only by the volume of information loss due to masking but also by the diversity of sentences resulting from random masking. However, the employment of ensemble techniques, even with baseline ensemble (vanilla prompts, not superior prompts), results in a more gradual increase in variance. This stabilization is likely due to the ensemble's ability to aggregate insights from multiple prompts, delivering a more consistent and reliable output despite the information loss introduced by masking. The superior prompt ensemble technique further reduces the variances.

\begin{figure}[ht!]
    \centering
    \includegraphics[width=1\linewidth]{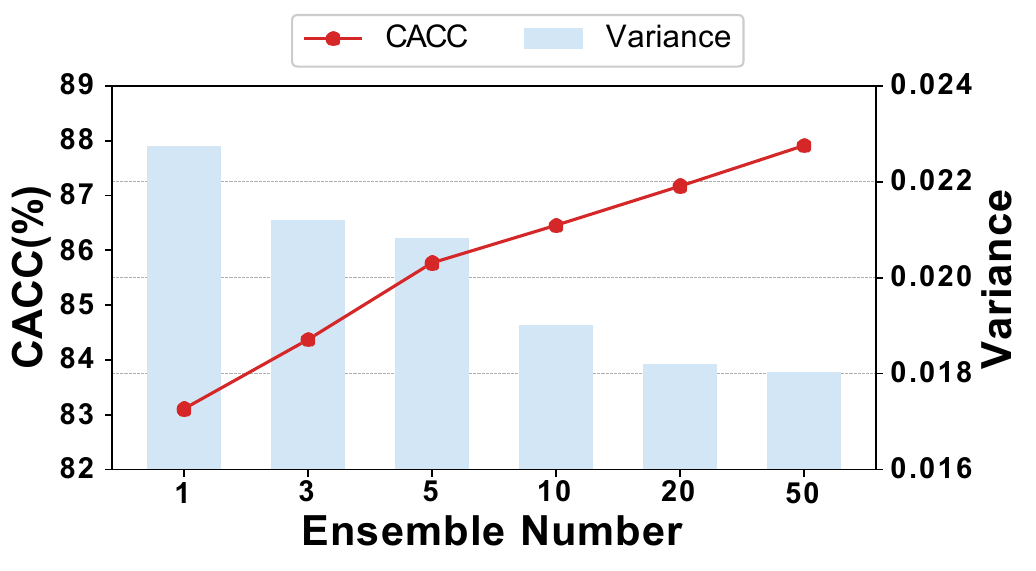}
    \caption{Clean accuracy and variance of CR-UTP under different ensemble numbers.}
    \label{fig:res_ens_num}
\end{figure}

\noindent\textbf{Ensemble Numbers.} To investigate the impact of the number of prompts within the superior prompt ensemble on clean accuracy, we conducted experiments on the SST-2 dataset using a large mask ratio of $70\%$ to amplify the ensemble number's effect on output performance. To mitigate the potential impact of differences in prompt selection performance on the output, each ensemble was selected to have similar mean accuracy. The results depicted in Figure~\ref{fig:res_ens_num} show that as the number of prompts in the ensemble increases from $1$ to $50$, there is a consistent improvement in clean accuracy, rising from $83.21\%$ to $87.82\%$, accompanied by a corresponding decrease in variance. These findings indicate that a larger ensemble leads to more stable and accurate predictions. This enhancement can be attributed to the ensemble's capacity to integrate diverse insights from multiple prompts, reducing the impact of any single erroneous prediction and fostering a consensus that is more resilient to the introduction of masks.

% \begin{table*}[ht!]
% \centering
% \footnotesize
% \setlength{\tabcolsep}{4pt}
% \caption{Compare with Random Mask~\cite{zeng2023certified}}
% \begin{tabular}{llllllllll}\toprule
%  & \multicolumn{3}{c}{TrojLLM} & \multicolumn{3}{c}{TextFool} & \multicolumn{3}{c}{DeepWordBug} \\\cmidrule(lr){2-4}\cmidrule(lr){5-7}\cmidrule(lr){8-10}
%  & CACC(\%) & ASR(\%) & PACC(\%) & CACC(\%) & ASR(\%) & PACC(\%) & CACC(\%) & ASR(\%) & PACC(\%) \\\midrule
% Random Mask & $69.74$ & $85.31$ & $56.84$ & $69.74$ & $37.88$ & $24.12$ & $69.74$ & $66.67$ & $33.33$ \\
% CR-UTP & $85.61$ & $50.55$ & $73.04$ & $85.61$ & $45.18$ & $54.82$ & $85.61$ & $37.39$ & $62.61$ \\\bottomrule
% \end{tabular}
% \label{t:comprasion_with_prior_work}
% \end{table*}

\begin{table}[ht!]
\centering
\footnotesize
\setlength{\tabcolsep}{2pt}
\caption{Evaluation on large language models.}
\begin{tabular}{lcccccc}\toprule
\multirow{2}{*}{Model} & \multicolumn{3}{c}{Llama2-7b} & \multicolumn{3}{c}{GPT-3.5} \\\cmidrule(lr){2-4}\cmidrule(lr){5-7}
 & CACC & ASR & PACC & CACC & ASR & PACC \\\midrule
  w/o defense &  $90.40$ &   $88.17$  & $53.82$ & $92.01$ &  $96.88$  & $51.94$ \\
Our baseline & $73.89$ &   $83.14$  & $55.03$ & $75.34$ &  $86.72$ & $56.19$ \\
    CR-UTP    & $84.68$ &   $51.47$   & $72.95$ & $85.32$ &  $50.75$ & $74.86$ \\\bottomrule
\end{tabular}

\label{table:differentModel}
\end{table}

\noindent\textbf{Evaluation on Large Language Models.} We demonstrate the effectiveness of our CR-UTP method on large models like Llama2-7b and GPT-3.5. The experiments, conducted on the SST-2 dataset against the UTP attack TrojLLM, are shown in Table \ref{table:differentModel}. For both the Llama2-7b and GPT-3.5 models, our CR-UTP approach improves the CACC by over 10\% compared to our baseline, while also achieving a reduction in ASR of more than 30\%.
 %However, when facing mask operations with a high mask ratio, these large models suffer a significant decline in performance. This is because BERT models are pre-trained using a Masked Language Model (MLM) approach, which closely resembles random masking. In contrast, GPT-3 and LLaMA2 models utilize the next token prediction method for their pre-training.

% \textbf{Performance on longer dataset}
% The table \ref{table:longer datatest} which shares the same settings as Table \ref{t:ablation_techniques}, demonstrates that our CR-UTP method is also effective on the longer dataset such as Yelp dataset \cite{DBLP:journals/corr/Asghar16}, with a consistent trend as those presented in Table \ref{t:ablation_techniques}. Our CR-UTP method showcases significant improvements over previous defenses. Specifically, on the Yelp dataset, it enhances the CACC and the PACC by 9.52\% and 13.5\%, respectively. Additionally, it reduces the ASR by 10.35\%. The results also demonstrate that the performance drop on datasets with longer text lengths is less significant than on datasets with shorter texts. This occurs because there is more remaining information after a large masking operation in longer text data.

% \begin{table}[ht!]
% \centering
% \footnotesize
% \setlength{\tabcolsep}{4pt}
% \begin{tabular}{lccc}\toprule
%  & CACC & ASR & PACC  \\\midrule
%   w/o Defense & 95.42 & 91.66   & 51.42 \\
% Baseline & 76.10 &   67.18   & 69.35 \\ 
% CR-UTP    &  85.62 &  56.83  & 82.85   \\\bottomrule
% \end{tabular}
% \caption{Performance on Yelp Dataset}
% \label{table:longer datatest}
% \end{table}

\begin{table}[ht!]
\centering
\footnotesize
\setlength{\tabcolsep}{2pt}
\caption{Comparison with adversarial training.}
\begin{tabular}{lcccccc}\toprule
\multirow{2}{*}{Attack} & \multicolumn{3}{c}{TrojLLM} & \multicolumn{3}{c}{TextFooler} \\\cmidrule(lr){2-4}\cmidrule(lr){5-7}
 & CACC & ASR & PACC & CACC & ASR & PACC \\\midrule
w/o defense &  $92.69$ & $91.88$  & $53.76$ & $92.69$ &   $92.27$  &  $8.13$\\
Adv. training &  $85.94$ &  $80.68$  & $59.82$ & $85.94$ &   $91.13$  &  $8.81$ \\
CR-UTP &  $85.70$ &  $50.55$  & $73.04$ & $85.28$&   $37.39$ & $62.61$ \\\bottomrule
\end{tabular}
\label{table:otherDefence}
\end{table}
\noindent\textbf{Comparison with Adversarial Training.}
We compare the empirical defence effects of adversarial training~\cite{yoo2021towards} and our CR-UTP in Table \ref{table:otherDefence}. Our CR-UTP significantly reduces the ASR by over 30\% while maintaining similar accuracy, outperforming adversarial training. CR-UTP consistently defends against various adversarial attacks, such as TrojLLM and TextFooler, unlike adversarial training, which shows inconsistent defense effectiveness. For instance, adversarial training effectively reduces ASR from 91.88\% to 80.68\% for attacks it was trained against TrojLLM, but it provides minimal defense against different attacks (TextFooler), only reducing ASR from 92.27\% to 91.13\%.

\section{Limitation}
The limitations of our paper are as follows:
(i) Certified Accuracy. Although our CR-UTR has demonstrated improvements in certified accuracy and reduced ASR, achieving state-of-the-art results, there remains a gap between clean accuracy and certified accuracy. (ii) Broader Applications. While our CR-UTP primarily concentrates on classification tasks, broadening its application to encompass other NLP tasks like generation~\cite{xue2024badrag} would present a captivating expansion of our research. (iii) Efficiency. The CR-UTP with prompt ensemble would result in higher inference overhead compared to using just one prompt. However, it is important to note that a superior prompt can significantly enhance the effectiveness of defense strategies. Moreover, the Superior Prompt Search is an offline process to train the policy model which could be reused to generate multiple superior prompts in several seconds.

%The CR-UTP with prompt ensemble could bring inference overhead than inference with one prompt. However, the CR-UTP would improve the performance against masking operation without the need to re-train the model. Meanwhile, the Superior Prompt Search is an offline process to train the policy model which could be reused to generate multiple superior prompts in several seconds. We underscore the efficiency of prompt ensembling over model ensembling, owing to the prompt generation's speed and low memory footprint, as opposed to the more resource-intensive generation and memory demands of model ensembling.

%(ii) Model. While our CR-UTP has been evaluated on popular benchmark datasets SST-2 and AgNews using the RoBERTa-large model, it would be beneficial to assess its effectiveness across more models and different architectures to ensure its generalizability.

\section{Conclusion}
In this paper, we address the challenge of certifying language model robustness against Universal Text Perturbations (UTPs) and input-specific text perturbations (ISTPs). We introduce the \textit{superior prompt search} method and the \textit{superior prompt ensembling} technique to enhance certified accuracy against UTPs and ISTPs. Our approaches achieve state-of-the-art results, ensuring stability and reliability in language model predictions across diverse attack scenarios.

\bibliography{ACL}
\newpage
\appendix
\section{Appendix}
\label{sec:appendix}
\subsection{Training Time and Inference Efficiency}
Regarding the training time, the Superior Prompt Search is an offline process before online inference, thus the prompt search and generation phase does not impact the online inference.  As mentioned in Section~\ref{sec:experiments}, the prompt search normally takes about 3.8 hours to train the policy model using only one single Nvidia GeForce RTX-3090 GPU. Once the policy model is trained, one could reuse it to generate multiple superior prompts in several seconds. The superior prompt is short and effective, comprising up to 5 tokens, resulting in an overhead (from appending the prompt compared to not using one) of no more than 10\% for both the SST and AgNews datasets. Additionally, datasets with longer texts (Yelp) exhibit much lower overhead ratios, i.e., less than 5\%. We highlight that superior prompt significantly enhances defense effectiveness. As the Table \ref{t:ablation_techniques} shows, on the SST-2 dataset, it yields an improvement of over 15\% in clean accuracy and a reduction of more than 30\% in ASR compared to our baseline.

The generation of ensemble prompts runs parallel to the search for the superior prompt, occurring before the online inference phase. The certified efficiency of this method is closely linked to the number of inference executions, which is a product ($kn$) of the ensemble number ($k$) and the sampling number ($n$). To maintain efficiency, one could reduce the sampling number ($n$) when employing ensemble prompts ($k>1$) to keep a similar or the same product $kn$. For instance, sampling a single superior prompt 5000 times may yield a certified accuracy of 54.5\%; however, an ensemble of 5 superior prompts (k=5) requires only 1000 samplings to reach a certified accuracy of 58.1\%.  For the defense inference against a specific attack, the number of $n$ can be much smaller, e.g., 50-100, for efficiency consideration. Also, We underscore the efficiency of prompt ensembling over model ensembling, owing to the prompt generation's speed and low memory footprint, as opposed to the more resource-intensive generation and memory demands of model ensembling.

\end{document}